\pdfoutput=1

\documentclass[11pt]{article}
\usepackage{ACL2023}

\usepackage{times}
\usepackage{latexsym}
\usepackage{multirow}
\usepackage{booktabs}
\usepackage{amsmath}
\usepackage{amssymb}
\usepackage{adjustbox}
\usepackage{amsfonts}

\usepackage{cleveref}
\usepackage{graphicx}
\usepackage{epstopdf}
\usepackage{hyperref}
\usepackage{colortbl}
\usepackage{float}
\usepackage{pifont}
\usepackage{xcolor}
\usepackage{mdframed}

\usepackage{tabularx}

\definecolor{my_green}{RGB}{51,102,0}
\definecolor{my_red}{RGB}{204, 0, 0}

\makeatletter
\let\scshape\relax 
\DeclareRobustCommand\scshape{%
  \not@math@alphabet\scshape\relax
  \ifnum\pdf@strcmp{\f@family}{\familydefault}=\z@
    \fontfamily{qbk}%
  \fi
  \fontshape\scdefault\selectfont}
\makeatother

\usepackage[T1]{fontenc}
\usepackage[utf8]{inputenc}
\usepackage{microtype}
\usepackage{inconsolata}
\usepackage{graphicx}

\title{Does RAG Really Perform Bad In Long-Context Processing?}  

\author{
    Kun Luo$^{1,2,*}$, \ Zheng Liu$^{1,4}$\thanks{ ~ Equal contribution, with Zheng Liu as the project lead.}, \ Peitian Zhang$^{1,3}$, \ Hongjin Qian$^{1,5}$, \ Jun Zhao$^{2}$, \ Kang Liu$^{2}$ \\ 
    $^1$  BAAI, \ \ \ \  $^2$  CASIA , \ \ \ \ $^3$ HKPU, \ \ \ \ $^4$  RUC, \ \ \ \ $^5$  PKU \\
    \texttt{\{luokun695,zhengliu1026,namespace.pt,chienqhj\}@gmail.com} 
}

\begin{document}
\maketitle
\begin{abstract}
The efficient processing of long context poses a serious challenge for large language models (LLMs). Recently, retrieval-augmented generation (RAG) has emerged as a promising strategy for this problem, as it enables LLMs to make selective use of the long context for efficient computation. However, existing RAG approaches lag behind other long-context processing methods due to inherent limitations on inaccurate retrieval and fragmented contexts. 
To address these challenges, we introduce \textbf{RetroLM}, a novel RAG framework for long-context processing. Unlike traditional methods, \textit{RetroLM employs KV-level retrieval augmentation}, where it partitions the LLM's KV cache into contiguous pages and retrieves the most crucial ones for efficient computation.
This approach enhances robustness to retrieval inaccuracy, facilitates effective utilization of fragmented contexts, and saves the cost from repeated computation. Building on this framework, we further develop a specialized retriever for precise retrieval of critical pages and conduct unsupervised post-training to optimize the model’s ability to leverage retrieved information. 
We conduct comprehensive evaluations with a variety of benchmarks, including LongBench, InfiniteBench, and RULER, where RetroLM significantly outperforms existing long-context LLMs and efficient long-context processing methods, particularly in tasks requiring intensive reasoning or extremely long-context comprehension. 


\end{abstract}

\section{Introduction}

The processing of long contexts has emerged as a critical issue in the development and application of Large Language Models (LLMs). Numerous applications necessitate the ability to handle extended sequences of information, including understanding lengthy documents \citep{bai2023longbench, caciularu2023peek}, supporting sophisticated AI agent systems \citep{jin2024long}, and generating long-form reasoning chains for complex tasks, such as mathematical proofs \citep{openai2024learning} or computer programming \citep{gur2023real}. To address this crucial requirement, substantial efforts have been devoted to extending the maximum context lengths accommodated by LLMs. For example, GPT-4 \citep{achiam2023gpt} and LLaMA-3.1 \citep{dubey2024llama}, both of which support a 128K token context window. Moreover, the recent Gemini-1.5-pro \citep{team2024gemini} makes a dramatic extension, enabling a context window of over 10M input tokens. 

Despite these advancements, the naive extension of context lengths remains constrained in several aspects. One significant challenge is the dramatic rise in computation when processing long contexts. As such, efficient long-context processing techniques have attracted growing interest. For example, StreamingLLM and LM-Infinite \citep{streamingllm,lminfinite} maintain the most recent KVs within a sliding window alongside initial attention sinks; while SnapKV and InfLLM \cite{snapkv,infllm} identify critical attention features for KV compression. Recently, retrieval-augmented generation (\text{RAG}) has emerged as a promising strategy for this problem \cite{xu2023retrieval,li2024long}. These approaches leverage retrievers to extract useful context fragments from very long inputs, which effectively overcomes the limits of LLMs' context lengths. By making selective use of the retrieved fragments, RAG further enables more efficient computation for long-context tasks. 

However, RAG-based methods are subject to the following inherent limitations while handling long-context tasks.  1) \textit{Retrieval Inaccuracy}. Many long-context processing tasks provide no explicit queries at all, like document summarization, code completion, and in-context learning \cite{bai2023longbench}. As a result, traditional retrievers become inapplicable to handle corresponding problems. Besides, it's non-trivial to properly chunk long contexts for retrieval \cite{qian2024grounding}, and it's hard for retrievers to deal with zero-shot settings. Without precise and complete acquisition of useful information, LLMs will be unable to produce correct outputs through RAG. 
2) \textit{Fragmented Contexts}. The retrieval operation introduces fragmented token spans from the input data, which are incoherent and prone to incompleteness. This significantly prevents LLMs from making effective use of the contextual information. 3) \textit{Repeated Computation}. The pre-filling operation needs to be re-conducted for the retrieved tokens of each task, both for retriever and generator, resulting in a huge waste of computation. \textit{Because of the above problems, existing RAG-based methods fall behind long-context LLMs and other efficient long-context processing approaches in many popular evaluation benchmarks} \cite{bai2023longbench,xu2023retrieval,li2024long}. 

In this paper, we propose \textbf{RetroLM}, a novel RAG framework designed for efficient long-context processing. RetroLM partitions the LLM's KV cache into contiguous pages and offloads them to external storage. During both pre-filling and decoding stages, it retrieves only the most crucial pages for the current context window, enabling efficient long-context processing. Unlike traditional RAG approaches which operate on raw tokens, retrieval augmentation at the KV cache level offers several advantages. First, it is robust to retrieval inaccuracy, as useful information within a certain token span can be captured by all succeeding KVs. Second, LLMs can naturally accommodate fragmented KVs due to the inherent sparsity of LLMs' attention patterns \cite{jiang2024minference}. Third, the KV cache is computed once and reused, thus eliminating repeated computation \cite{pope2023efficiently}. 

We introduce a couple of key operations to optimize the performance of RetroLM. For precise retrieval of crucial pages, we design a specialized \textbf{page retriever}. It estimates the pages' importance using fine-grained KV interactions; and by fine-tuning over well-curated datasets, it achieves strong generality across various downstream tasks and a broad scope of context lengths. To make better use of fragmented KVs, we perform \textbf{post-training} based on unlabeled data. This further contributes to the end-to-end performance of RetroLM. 

We perform comprehensive evaluations using several standard benchmarks in this field, including LongBench \cite{bai2023longbench}, InfiniteBench \cite{infbench}, and RULER \cite{hsieh2024ruler}. In our experiment, RetroLM outperforms popular efficient long-context processing methods with notable advantages. In majority of the tasks, it achieves an equivalent performance as the expensive full-attention methods; while for certain scenarios like long-doc QA, it even surpasses full-attention by effectively filtering out background
noise and focusing on the most useful KV entries. Our well-trained models and source code will be made publicly available to facilitate future research.

\section{Related Work}
In this section, we make discussions on the following related works: 1) context extension of LLMs, 2) efficient long-context processing, 3) RAG approaches for long-context processing.  

First of all, a substantial body of research has focused on extending the context length of LLMs directly. One common approach involves modifying positional encoding mechanisms to enable LLMs trained on short texts to process longer inputs directly during inference \citep{chen2023extending, peng2023yarn, ding2024longrope}. While straightforward, these methods often yield suboptimal performance without additional fine-tuning. Another widely adopted strategy is continual training, where existing LLMs are fine-tuned on long-sequence data to expand their context windows \citep{li2023long, chen2023longlora, mohtashami2023landmark, xiong2023effective}. However, fine-tuning approaches typically require training from extremely long-sequence data, which is challenging due to the scarcity of native human-annotation data and the high expenses resulted from the training operations \citep{fu2024data, gao2024train}. 


Recent studies have explored various types of efficient long-context processing techniques to alleviate computational and memory constraints \citep{sun2024shadowkv, liao2024e2llm, yang2024pyramidinfer}.
Stream processing approaches, such as StreamingLLM \citep{streamingllm} and LM-Infinite \citep{lminfinite}, maintain the most recent KVs within a sliding window alongside initial attention sinks. Sequential compression techniques, such as Activation Beacon \citep{zhang2024soaring}, compress intermediate activations into more compact forms to conserve memory. KV quantization methods, including KIVI \citep{liu2024kivi}, encode the KV cache using low-bit representations to minimize storage requirements. 
Among these methods, KV cache sparsification has gained significant attention for their ability to selectively utilize portions of KVs based on certain reduction strategies, where KVs are reduced into a fixed budget (e.g., 2K) \citep{xu2024think, quest, huang2024locret, liu2024retrievalattention, shi2024discovering}. For instance, InfLLM \citep{infllm} incorporates intermediate information by segmenting KVs into fixed-size chunks and selecting top-k most salient chunks based on attention score patterns. H2O \cite{h2o} introduces a policy that greedily drops KVs during generation using a scoring function derived from cumulative attention. SnapKV and PyramidKV \citep{snapkv, pyramidkv} extend to alleviate memory pressure during the prefilling stage by dropping tokens based on cumulative attention scores within localized windows. 

Retrieval-augmented generation (RAG) has emerged as a promising approach for addressing long-context tasks \citep{xu2023retrieval, li2024long, yue2024inference}. Leveraging modern dense retrievers \citep{karpukhin2020dense, xiao2023c}, these approaches first partition the long text into smaller chunks, subsequently selecting the most salient chunks, and concatenating them to form a new prompt for the LLM \citep{zhao2024retrieval}. In addition, several specialized retrievers have been developed for long-context scenarios \citep{luo2024bge, gunther2023jina}. In this work, RetroLM integrates retrieval augmentation directly at the KV cache level, thereby seamlessly incorporating RAG pipeline into long-context language modeling.

\section{Method}

\begin{figure*}[t]
\centering
\includegraphics[width=0.9\linewidth]{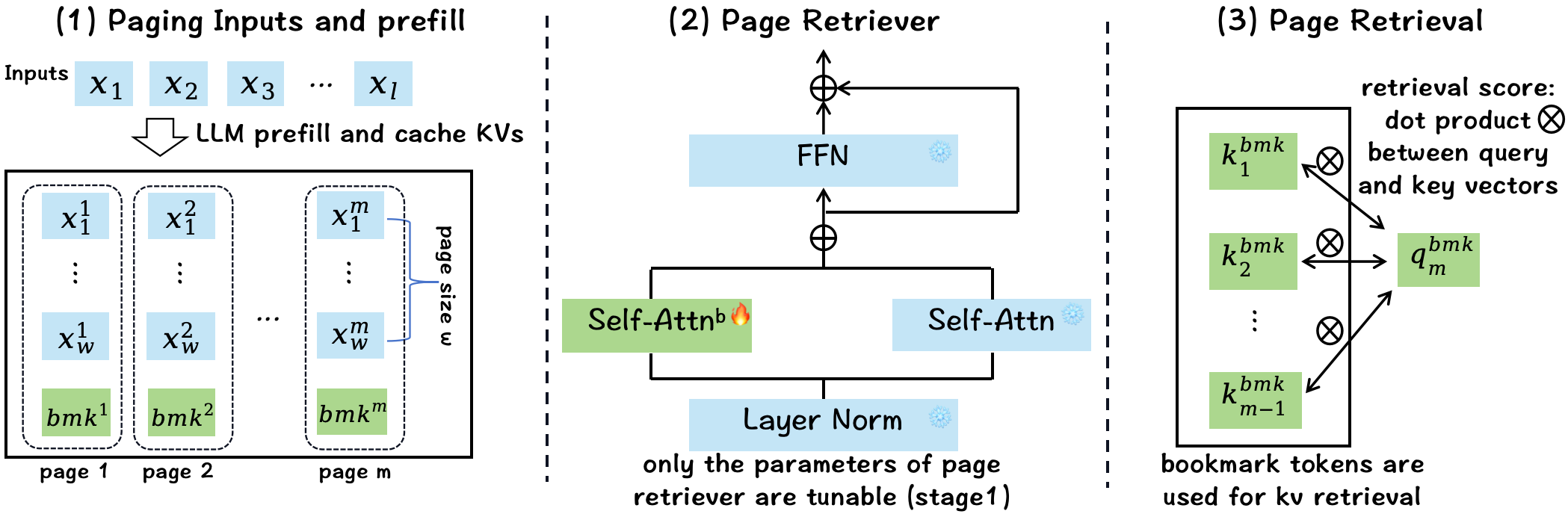}
\vspace{-12pt}
\caption{\small Framework of RetroLM: \textit{(1)} Paging mechanism for KV management. \textit{(2)} Specialized trainable, plug-in page retriever. \textit{(3)} KV retrieval using special bookmark tokens, with their representation established within attention module.}
\label{fig:framework}
\vspace{-12pt}
\end{figure*}

\subsection{Problem Formulation}
For long-context understanding and language modeling tasks, such as question answering, summarization, the input can be structured into: context $X$, user query $q$, and target output $Y$. The generation objective of LLM can be expressed as: 
\begin{equation}
    max.\log\mathrm{LLM}(y_t|X,q,Y_{<t})
\end{equation}
In such scenarios, the context $X$ often exceeds 100K tokens, leading to significant computational and memory consumption. To address this problem, various efficient long-context processing techniques have been introduced, aiming at compressing context either implicitly or explicitly using a designated reduction policy $p(X)$.

RAG-based methods employ a standalone retriever as an explicit context reduction policy $p^{ret}(\cdot)$. It first chunks the long context into: $X:\{s_1, ..., s_N\}$, and then select the top-$k$ relevant chunks: $ X^{ret}:\{s_1, ..., s_k\}$. The $X^{ret}$ forms the new input context. Explicit context compression of RAG-based methods prune the prompt rigidly, which results in information loss and semantic discontinuities.

RetroLM performs retrieval augmentation at KV cache level, using a plug-in page retriever as policy $p^{kv}(\cdot)$. 
It selects the most crucial KVs at each decoder layer: $C=p^{kv}(X)$, where $C$ is the KVs for attention computation, thereby achieving implicit context compression. 
Unlike existing KV sparsification approaches that rely on heuristic methods to approximate full attention, RetroLM introduces a specialized and trainable page retriever, inspired by dense retrieval techniques. We conduct further analysis in Sec. \ref{Sec:Case Study} to demonstrate the effectiveness of page retriever over full attention.

\subsection{Inference Process}
\noindent \textbf{Paging Inputs.} RetroLM first partitions the LLM's input context $X= \{x_i\}_{i=1}^{l}$ into contiguous pages: 
\begin{equation}
\small
\{x_1, ..., x_{l}\} \xrightarrow{\mathrm{partition}} \{X_1, ..., X_m\}, X_i = \{x_j^i\}_{j=1}^{w}
\end{equation}
where $w$ is the page size (128 in practice).
Then for each page $X_i$, a special bookmark token ( $\small \mathrm{\langle BMK \rangle}$) is inserted to the end of it: $\small X_i^{'} = \{x_1^i, ..., x_w^i, {\langle bmk \rangle}^i\}$. 
The LLM encodes both the normal tokens and bookmark tokens. The bookmark tokens function as the \textbf{\textit{page indexs}} of corresponding pages for KV retrieval and establish their representations during attention computation across each decoder layer. 

\noindent \textbf{Pre-filling.} During pre-filling, we employ streaming encoding based on page retrieval to enable the process of extremely long inputs.
Specifically, a fixed-sized sliding window is used to encode the long context progressively.
In each layer, the encoding of page $X_{i}{'}$ only retrieves $k$ pages (including the first page as attention sink) for attention computation instead of costly full attention: 
\begin{equation}
\small
C: \{X_1', ..., X_k'\} = p^{kv}(X':\{X_1', ..., X_{i-1}'\}|X_i')
\end{equation}
Once encoded, the KVs of page $X_i'$ are offloaded to CPU, ensuring that only the required KV pages are reloaded to GPU for attention computation.

\noindent \textbf{Decoding.} During decoding, page retrieval is conducted only once given the user query: 
\begin{equation}
\small
C: \{X_1', ..., X_k'\} = p^{kv}(X':\{X_1', ..., X_{m}'\}|q)
\end{equation}

\subsection{Page Retriever}
\textbf{Architecture.} We propose a trainable, plug-and-play page retriever designed to conduct KV cache level retrieval augmentation, whose architecture is shown in Figure \ref{fig:framework} (Middle). It reuses all modules of the LLM except imposing a slight modification on the self-attention module. 

During the self-attention computation, the hidden states of normal tokens ($n$) and bookmark tokens ($b$) are sliced out and projected into query, key, and value vectors respectively: 
\begin{equation}
\small
\begin{aligned}
    \boldsymbol{Q}^n &= \boldsymbol{W}_Q^n \boldsymbol{H}^n, \quad 
    \boldsymbol{K}^n = \boldsymbol{W}_K^n \boldsymbol{H}^n, \quad 
    \boldsymbol{V}^n = \boldsymbol{W}_V^n \boldsymbol{H}^n, \\
    \boldsymbol{Q}^b &= \boldsymbol{W}_Q^{b} \boldsymbol{H}^{b}, \quad 
    \boldsymbol{K}^b = \boldsymbol{W}_K^{b} \boldsymbol{H}^{b}, \quad 
    \boldsymbol{V}^b = \boldsymbol{W}_V^{b} \boldsymbol{H}^{b}
\end{aligned}
\end{equation}
where $\small\boldsymbol{W_*^n}$ are the LLM’s original projection matrices and $\small\boldsymbol{W_*^{b}}$ are the newly introduced matrices designed specifically to handle bookmark tokens. The bookmark tokens distill corresponding page's contextual information during attention computation and are used for page retrieval.

\noindent \textbf{Retrieval Score.} Page importance estimation employs similarity between the query vector of target page's bookmark token and the key vectors of past pages' bookmark tokens: 
\begin{equation}
\small
p^{kv}(\{X_1', ..., X_{m-1}'\}|X_m') = \text{top-}k \left\{\langle \boldsymbol{q}_m^{bmk}, \boldsymbol{k}_j^{bmk} \rangle \right\}_{j=1}^{m-1}
\end{equation}
where $\langle *, * \rangle$ denotes the dot product operation, commonly used as a similarity measurement in dense retrieval \citep{karpukhin2020dense}.

\noindent \textbf{Training.} Training the page retriever poses a challenge due to the lack of appropriately labeled long-context data for retrieval supervision signals. Drawn inspiration from the training paradigm of advanced dense retrievers, where a few negative samples are employed to establish the ability to distinguish relevant passages from corpus containing millions of samples, we adopt contrastive learning to train the page retriever for strong generalizability and robustness \citep{karpukhin2020dense, chen2024bge, luo2024bge}.

As shown in Figure \ref{fig:data_stage1_msm}, we leverage 50K pairwise data from the dense retrieval training set MS MARCO \cite{bajaj2016ms}, which is derived from real-world web search queries, to provide valuable transferable semantic matching capabilities for the page retriever. To construct input sequences, we concatenate the positive passage with hard negative passages in a random order, forming pseudo-texts up to a length of 8K tokens, and append the web search query to the end of the sequence (page $X_m'$). Additionally, we synthesize 5K pairwise samples using text from Slimpajama \citep{shen2023slimpajama}, which contains coherent contexts that enable page retriever to effectively learn to find target KVs relevant to query. The detailed data format and training implementation are described in Appendix \ref{Appendix: Training Detail and Data Formulation}. 

Assuming the useful KVs for the local page $m$ (where the query resides) is located on page $i$ (denoted as $X_i'$) the contrastive learning objective is defined as follows:
\begin{equation}
\small
L_{1} = \large -\log\frac{\exp(\langle \boldsymbol{q}_m^{bmk},\boldsymbol{k}_{i}^{bmk} \rangle)}{\sum_{j=1}^{m-1}\exp(\langle\boldsymbol{q}_m^{bmk},\boldsymbol{k}_j^{bmk}\rangle)}
\end{equation}
where $\boldsymbol{q}_*^{bmk}$ and $\boldsymbol{k}_*^{bmk}$ are the query and key vectors of bookmark tokens of corresponding pages in the self-attention module. This training phase, referred to as \textbf{Stage-1}, focuses on training the page retriever to identify useful KVs against complex and distracting contexts at each decoder layer, while keeping the backbone LLM frozen.

\subsection{Post Training}
We conduct \textbf{Stage-2} post-training for RetroLM, during which model parameters are fine-tuned to adapt to sparse KV caches retrieved by the page retriever. Our training leverages unsupervised pretraining data from SlimPajama \citep{shen2023slimpajama} (up to 12K tokens) and employs a streaming encoding strategy. During language modeling, each page uses the well-trained page retriever to select most semantically important top-$k$ pages for attention computation (similar to inference process), instead of relying on full attention: 
\begin{equation}
\small
C: \{X_1', ..., X_k'\} = p^{kv}(\{X_1', ..., X_{i-1}'\}|X_i')
\end{equation}
The loss function follows the standard language modeling loss formula:
\begin{equation}
L_{2} = -\sum_{t} \log P(x_t | x_{<t})
\end{equation}
The training data consists of unsupervised, relatively short-length corpus segments from SlimPajama \citep{shen2023slimpajama}, with a maximum token length of 12K. Notably, the objective is not length extension but rather enhancing the model's capacity for adaptation to retrieved sparse KV pages. We use the same data to finetune LLM directly for further analysis in Sec. \ref{Sec:Ablation}.

\section{Experiment} 
We conduct extensive experiments focused on answering the following two research questions: 1) The effectiveness of RetroLM against long-context LLMs and other efficient methods. 2) How well can RetroLM generalize to different long-context tasks and context lengths.

\subsection{Setting}
\textbf{Datasets.} To comprehensively evaluate the overall performance of RetroLM, we employ the LongBench suite \cite{bai2023longbench}. This benchmark encompasses a variety of tasks, including single-document QA, multi-hop QA, summarization, and long ICL. These tasks are well-suited for assessing the long-context capability of different methods in practical application scenarios. Subsequently, to assess the generalization of RetroLM in extremely long scenarios, we utilize several realistic and representative tasks from InfiniteBench \citep{infbench}, including free-form QA on long books (QA), summarization over long texts (Summary), multiple-choice QA on long books (Choice), and finding special numbers in lengthy lists (Math.F). The average input length within InfiniteBench is 145K tokens. We also use RULER \citep{hsieh2024ruler} to evaluate long context key information identification capability. All evaluation metrics are aligned with official implementation.

\noindent \textbf{Baseline Methods.} To rigorously demonstrate the effectiveness of RetroLM, we compare its performance against the following competitive baseline methods: (1) Original Models: We report the performance of the LLMs with full attention mechanisms \citep{jiang2023mistral, dubey2024llama}. (2) Stream Processing: This category includes methods like LM-Infinite \citep{lminfinite} and StreamingLLM \citep{streamingllm}, which employ attention sink and sliding window mechanisms for processing long inputs. (3) KV Sparsification: These methods, such as H2O \citep{h2o}, SnapKV \citep{snapkv}, InfLLM \citep{infllm}, and PyramidKV \citep{pyramidkv}, employ heuristic KV sparsification policies to selectively retain portions of KVs. (4) RAG: We employ several retrieval methods to conduct RAG pipeline: the classic BM25 method \citep{robertson2009probabilistic}, the Contriever model \citep{izacard2021unsupervised}, and the SOTA BGE-large-v1.5 model \citep{xiao2023c}.

\begin{table*}[htbp]
\centering
\vspace{-10pt}
\makebox[\textwidth][c]{
\resizebox{\textwidth}{!}{
\begin{tabular}{lcccccccccccccc}
\toprule
\textbf{Model} & \textbf{Context} & \textbf{Narrative} & \textbf{Qasper} & \textbf{Multifield} & \textbf{Hotpot} & \textbf{2wikim} & \textbf{Musique} & \textbf{GovReport} & \textbf{MultiNews} & \textbf{QmSum} & \textbf{Trec} & \textbf{Trivia} & \textbf{SAMSum} & \textbf{Average} \\
\midrule
\multicolumn{15}{c}{\textbf{Mistral-7B-Instruct-v0.2}} \\
\midrule
Mistral-7B-v0.2 & 32k & 26.9 & 33.1 & 49.2 & 43.0 & 27.3 & 18.8 & 25.6 & 26.2 & 23.3 & 71.0 & 86.2 & 42.6 & 39.4 \\
LM-Infinite & 2k & 20.4 & 26.9 & 45.1 & 36.1 & 24.2 & 14.0 & 27.1 & 24.3 & 21.6 & 68.0 & 72.2 & 31.7 & 34.3 \\
StreamingLLM & 2k & 20.3 & 26.6 & 45.7 & 35.3 & 24.3 & 12.2 & 27.5 & 24.5 & 21.6 & 68.5 & 71.9 & 31.2 & 34.1 \\
InfLLM & 2k & 23.5 & 28.8 & 47.7 & 41.3 & 25.7 & 17.5 & 29.1 & 26.3 & 21.2 & 68.0 & 84.4 & 41.4 & 37.9 \\
H2O & 2k & 25.6 & 31.1 & 49.0 & 40.8 & 26.5 & 17.1 & 24.8 & 26.6 & 23.6 & 55.0 & 86.3 & 42.4 & 37.4 \\
SnapKV & 2k & 25.9 & 32.9 & 48.6 & 43.0 & 27.4 & 19.0 & 26.6 & 26.7 & 24.4 & 70.0 & 86.2 & 42.5 & 39.4 \\
PyramidKV & 2k & 25.5 & 32.2 & 49.0 & 42.3 & 27.5 & 19.4 & 26.6 & 26.7 & 24.0 & 71.0 & 86.2 & 42.9 & 39.4 \\
RetroLM-Stage1 & 2k & 26.8 & 34.0 & 50.8 & 47.6 & 39.0 & 22.5 & 29.3 & 27.3 & 24.6 & 69.5 & 88.8 & 42.4 & 41.9 \\
RetroLM-Stage2 & 2k & 26.6 & 38.7 & 53.8 & 47.7 & 41.6 & 26.4 & 29.8 & 28.2 & 25.9 & 70.5 & 89.3 & 43.0 & 43.5 \\
\midrule
\multicolumn{15}{c}{\textbf{Llama-3-8B-Instruct}} \\
\midrule
Llama-3-8B & 8k & 25.8 & 29.6 & 41.0 & 45.4 & 36.1 & 22.9 & 26.2 & 26.5 & 23.4 & 74.0 & 90.5 & 42.3 & 40.3 \\
LM-Infinite & 2k & 22.0 & 26.2 & 38.3 & 40.5 & 33.1 & 17.1 & 23.0 & 26.5 & 22.5 & 70.0 & 83.1 & 32.2 & 36.2 \\
StreamingLLM & 2k & 21.7 & 25.8 & 38.1 & 40.1 & 32.0 & 16.9 & 23.1 & 26.5 & 22.6 & 70.0 & 83.2 & 31.8 & 36.0 \\
InfLLM & 2k & 23.4 & 29.0 & 40.9 & 41.5 & 34.3 & 19.7 & 25.7 & 26.8 & 22.4 & 73.0 & 89.9 & 41.3 & 39.0 \\
H2O & 2k & 25.6 & 26.9 & 39.5 & 44.3 & 32.9 & 21.1 & 24.7 & 24.6 & 23.0 & 53.0 & 90.5 & 41.8 & 37.3 \\
SnapKV & 2k & 25.9 & 29.6 & 41.1 & 45.0 & 35.8 & 21.8 & 26.0 & 26.5 & 23.4 & 73.5 & 90.5 & 41.6 & 40.1 \\
PyramidKV & 2k & 25.4 & 29.7 & 40.3 & 44.8 & 35.3 & 22.0 & 26.8 & 26.2 & 23.3 & 73.0 & 90.5 & 42.1 & 40.0 \\
RetroLM-Stage1 & 2k & 25.4 & 33.8 & 48.7 & 50.2 & 39.8 & 24.1 & 26.9 & 27.0 & 24.7 & 73.5 & 91.0 & 42.2 & 42.3 \\
RetroLM-Stage2 & 2k & 26.6 & 38.7 & 48.9 & 52.5 & 45.4 & 27.0 & 30.4 & 27.9 & 26.1 & 75.5 & 90.7 & 42.8 & 44.4 \\
\bottomrule
\vspace{-17pt}
\end{tabular}
}
}
\caption{Experiment results of comparing RetroLM with other efficient processing methods on LongBench. The result emphasizes the effectiveness of RetroLM over strong baselines and a wide variety of tasks.}
\label{tab:longbench}
\end{table*}

\begin{table*}[ht]
\centering
\small
\setlength{\tabcolsep}{4pt}
\renewcommand{\arraystretch}{1.05}
\begin{tabular*}{0.99\textwidth}{@{\extracolsep{\fill}}lcccccccc@{}}
    \toprule
    \textbf{Model} & \textbf{Context} & \textbf{Narrative} & \textbf{Qasper} & \textbf{Multifield} & \textbf{Hotpot} & \textbf{2wikim} & \textbf{Musique} & \textbf{Average} \\
    \midrule
    Mistral-7B-v0.2 & 32k & 26.9 & 33.1 & 49.2 & 43.0 & 27.3 & 18.8 & 33.1 \\
    Mistral-BM25 & 2k & 13.9 & 22.7 & 34.6 & 31.0 & 22.7 & 17.8 & 23.8 \\
    Mistral-Contriever & 2k & 20.8 & 30.7 & 47.2 & 35.7 & 30.1 & 18.2 & 30.4 \\
    Mistral-BGE & 2k & 22.4 & 31.2 & 47.8 & 37.9 & 30.6 & 18.5 & 31.4 \\
    RetroLM-Stage1 & 2k & 26.8 & 34.0 & 50.8 & 47.6 & 39.0 & 22.5 & 36.8 \\
    RetroLM-Stage2 & 2k & 26.6 & 38.7 & 53.8 & 47.7 & 41.6 & 26.4 & 39.1 \\
    \bottomrule
    \vspace{-12pt}
\end{tabular*}
\caption{Experiment results of comparing RetroLM with RAG methods on LongBench QA tasks.}
\label{tab:rag_kv_comparison}
\vspace{-12pt}
\end{table*}

\subsection{Comparing with other Efficient Processing Methods on LongBench}
\label{sec:LongBench}
We compare stage-1 and stage-2 training of RetroLM with other efficient processing methods, using two popular backbone LLM (\textbf{Mistral-7B-Instruct} and \textbf{Llama-3-8B-Instruct}). The results on LongBench \citep{bai2023longbench} are presented in Table \ref{tab:longbench}. For the original models, we evaluate using their maximum context lengths. For RetroLM and other baseline methods, a fixed KV budget of 2K tokens is employed. Consequently, in each decoder layer's attention module, 2K tokens are selected for attention computation according to each method's respective KV reduction policy.

For the Mistral-based models, RetroLM achieves an overall score that surpasses all baselines, also significantly outperforming results obtained using full attention. Other approaches that employ heuristic KV selection strategies encounter performance ceilings comparable to full attention. Notably, RetroLM exceeds the performance of full attention by 2.5 points, even when only the KV retriever is trained during stage-1, with the language model remaining frozen. By learning to discriminate key information during the training of page retriever, RetroLM effectively identifies important KVs within extensive texts, achieving significant performance gains under constrained token budgets.

During stage-2 training of RetroLM, additional adaptation of LLM on unsupervised text data yields further performance improvements across tasks. This demonstrates the model's ability to adapt effectively to sparse KV cache and streaming encoding paradigm. To validate and analyze these findings, we conducted ablation studies using the same data but trained and evaluated the models with full attention (see Sec. \ref{Sec:Ablation}). Similar trends are observed in experiments with the Llama-3-based models, corroborating the generality of our findings.

\begin{table*}[t]
\centering
\footnotesize
\setlength{\tabcolsep}{6pt} 
\renewcommand{\arraystretch}{1.2} 
\begin{tabular*}{0.99\textwidth}{@{\extracolsep{\fill}}lcccccc@{}}
    \toprule
    \textbf{Model} & \textbf{Context} & \textbf{QA} & \textbf{Summary} & \textbf{Choice} & \textbf{Math.F} & \textbf{Average} \\
    \midrule
    Mistral-7B-v0.2 & 32k & 12.9 & 25.9 & 44.5 & 20.6 & 25.9 \\
    StreamingLLM & 6k & 10.9 & 21.0 & 40.4 & 15.1 & 21.8 \\
    H2O & 6k & 14.2 & 23.7 & 43.7 & 24.2 & 26.5 \\
    InfLLM & 6k & 15.0 & 24.1 & 41.7 & 24.9 & 26.5 \\
    SnapKV & 6k & 16.2 & 25.3 & 44.0 & 24.7 & 27.5 \\
    RetroLM-Stage1 & 6k & 18.4 & 27.8 & 45.0 & 24.2 & 28.9 \\
    RetroLM-Stage2 & 6k & 20.2 & 29.2 & 46.1 & 24.5 & 30.0 \\
    \bottomrule
    \vspace{-10pt}
\end{tabular*}
\caption{Experiment results on InfiniteBench. The results demonstrate the effectiveness and generalization of RetroLM across ultra-long contexts compared with other efficient processing methods.}
\label{tab:infbench}
\vspace{-8pt}
\end{table*}

\subsection{Comparing with RAG on LongBench}
In this section, we compare RetroLM with retrieval-augmented generation (RAG) methods, which similarly aim to identify and utilize query-relevant information from long contexts. The experimental results on LongBench \citep{bai2023longbench} QA tasks are presented in Table \ref{tab:rag_kv_comparison}. For RAG, we retrieve the top 10 most similar chunks (each 200 tokens) for each dataset.

The results demonstrate that RetroLM consistently outperforms all RAG methods. These findings highlight the superior ability of RetroLM to effectively utilize long-context information, which can be attributed to its dynamic KV retrieval mechanism. Unlike RAG methods, which rely on a static selection of information at the input stage, RetroLM dynamically retrieves crucial KVs at each decoder layer. This dynamic approach enables RetroLM to preserve crucial information and maintain global contextual awareness. While RAG's rigid prompt selection often leads to the permanent loss of relevant information that the retriever fails to identify \citep{xu2023retrieval}. Moreover, RAG's method for handling long-context tasks necessitates additional retrieval models and stages, thereby increasing the complexity of the task flow. In contrast, using RetroLM for long-context tasks allows for an end-to-end approach using a single model.

RetroLM's training of KV retriever in Stage-1 draws inspiration from contrastive learning paradigms employed in general-purpose dense retriever training \citep{izacard2021unsupervised, ma2024fine}. This design allows it to accurately identify useful KV pages from complex, noisy contexts (hard negatives). As a result, RetroLM can be viewed as a novel model-based RAG framework, integrating retrieval functionality directly into the LLM at KV cache level. This integration enhances both semantic comprehension and portability. We believe this approach holds broader research value in the future, including extensions to more complex long-context reasoning scenarios and the development of knowledge bases capable of storing and retrieving KVs.

\subsection{Experiment Results on InfiniteBench}
\label{sec:InfiniteBench}
The experimental results on InfiniteBench \citep{infbench} are presented in Table \ref{tab:infbench}. We compare RetroLM with other efficient processing methods to demonstrate its effectiveness and generalization across ultra-long contexts. Given that the lengths of most evaluation cases exceed 100K, we allocated a larger KV budget of 6K for all baselines.

Across all tasks, RetroLM consistently outperforms the full-attention baseline. This indicates that RetroLM effectively generalizes in scenarios involving ultra-long texts, despite being trained on significantly shorter context lengths. Specifically, during Stage-1, the KV retriever was trained on contexts up to 8K tokens, while in Stage-2, the language model was trained with an unsupervised corpus, using a maximum context length of 12K tokens. 

When compared to other efficient processing methods, RetroLM demonstrates a clear performance advantage. In the lengthy QA and summarization tasks, RetroLM-Stage2 outperforms SnapKV by 4.0 and 3.9 points respectively. This underscores RetroLM's potential as a scalable and effective solution for real-world applications that require processing of extremely long text.

\begin{table}[t]
\centering
\vspace{3pt}
\resizebox{0.99\columnwidth}{!}{
\begin{tabular}{lccccc|c}  
\toprule
\textbf{Model} & \textbf{4K} & \textbf{8K} & \textbf{16K} & \textbf{32K} & \textbf{64K} & \textbf{AVG} \\
\midrule
\multicolumn{7}{c}{\textbf{NIAH Performance}} \\
\midrule
Mistral-7B-v0.2 & 98.1 & 96.2 & 94.3 & 85.5 & 51.1 & 85.4 \\
RetroLM-Stage2  & 99.1 & 96.4 & 92.2 & 88.6 & 79.0 & 91.1 \\
\midrule
\multicolumn{7}{c}{\textbf{GPU Memory (GB)}} \\
\midrule
Mistral-7B-v0.2 & 17.0G & 19.0G & 22.0G & 28.6G & 43.3G & - \\
RetroLM-Stage2  & 18.3G & 18.7G & 19.3G & 20.1G & 25.5G & - \\
\bottomrule
\end{tabular}
}
\caption{Experiment results of NIAH tasks on RULER and GPU memory usage at different input lengths.}
\label{tab:RULER}
\vspace{-8pt}
\end{table}

\subsection{Experiment Results on RULER}
Beyond downstream long-context understanding tasks such as QA and summarization, we assess long-context retrieval capability of RetroLM using eight Needle-in-a-Haystack tasks from RULER \citep{hsieh2024ruler}. These tasks cover a diverse range of needle types and quantities with varying levels of difficulty, requiring the model to extract relevant information from a vast number of distractors. As shown in Table \ref{tab:RULER} (Top), RetroLM achieves superior performance compared to the full-attention Mistral model across evaluation lengths ranging from 4K to 64K, demonstrating robust long-context information identification capability.

\begin{table*}[t]
    \centering
    \small
    \setlength{\tabcolsep}{4pt}
    \renewcommand{\arraystretch}{1.05}
    \begin{tabular*}{0.99\textwidth}{@{\extracolsep{\fill}}lcccccccc@{}}
        \toprule
        \textbf{Model} & \textbf{Context} & \textbf{Narrative} & \textbf{Qasper} & \textbf{Multifield} & \textbf{Hotpot} & \textbf{2wikim} & \textbf{Musique} & \textbf{Average} \\
        \midrule
        Mistral-7B-v0.2 & 32k & 26.9 & 33.1 & 49.2 & 43.0 & 27.3 & 18.8 & 33.1 \\
        \midrule
        \multicolumn{9}{c}{\textbf{Ablation Study}} \\
        \midrule
        RetroLM w/o Stage1 & 2k & 23.6 & 29.9 & 45.4 & 38.5 & 24.9 & 15.1 & 29.6 \\
        RetroLM-Stage1 & 2k & 26.8 & 34.0 & 50.8 & 47.6 & 39.0 & 22.5 & 36.8 \\
        RetroLM-Stage2 & 2k & 26.6 & 38.7 & 53.8 & 47.7 & 41.6 & 26.4 & 39.1 \\
        Mistral-Finetuned & 32k & 26.9 & 33.4 & 48.5 & 44.5 & 30.6 & 19.4 & 33.9 \\
        InfLLM-Finetuned & 2k & 25.4 & 30.7 & 48.0 & 43.7 & 29.2 & 18.0 & 32.5 \\
        \midrule
        \multicolumn{9}{c}{\textbf{Analytical Experiment with Varying Budgets}} \\
        \midrule
        SnapKV (1024) & 1024 & 25.4 & 29.5 & 49.0 & 40.9 & 25.7 & 18.3 & 31.5 \\
        SnapKV (2048) & 2048 & 25.9 & 32.9 & 48.6 & 43.0 & 27.4 & 19.0 & 32.8 \\
        RetroLM-Stage1 (512) & 512 & 25.0 & 30.4 & 47.0 & 42.9 & 30.4 & 17.9 & 32.3 \\
        RetroLM-Stage1 (1024) & 1024 & 25.4 & 31.5 & 47.9 & 45.4 & 33.7 & 21.1 & 34.2 \\
        RetroLM-Stage1 (2048) & 2048 & 26.8 & 34.0 & 50.8 & 47.6 & 39.0 & 22.5 & 36.8 \\
        \bottomrule
        \vspace{-14pt}
    \end{tabular*}
    \caption{Analytical experiments with QA tasks from LongBench.}
    \label{tab:analytical_experiment}
    \vspace{-11pt}
\end{table*}

\begin{figure}[t]
\centering
\resizebox{1.01\linewidth}{!}{ 
    \includegraphics{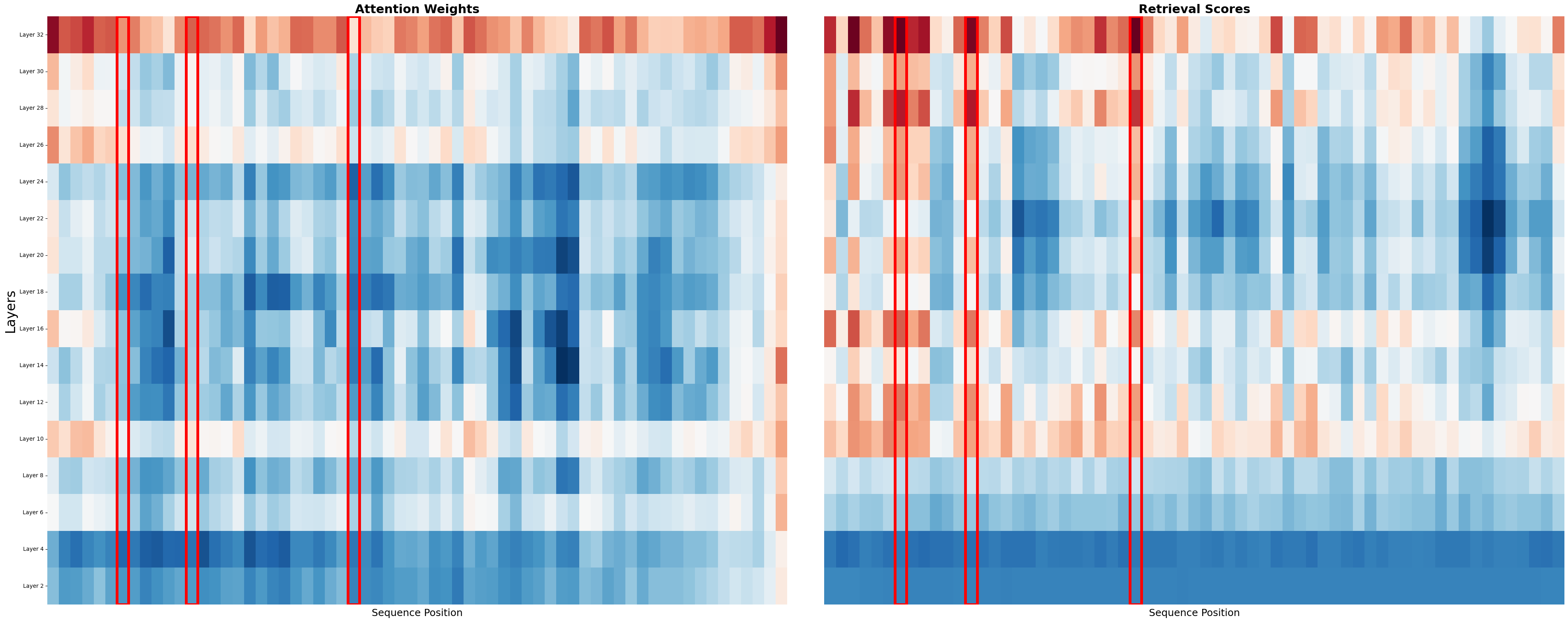}
}
\vspace{-18pt}
\caption{\small Attention score maps for a MusiQue case. Left: from original full attention. Right: score from RetroLM's KV retriever. \raisebox{0pt}{\scalebox{0.9}{\fcolorbox{red}{white}{Red squares}}} are answers for the multi-hop question. X-axis represents sequence position, Y-axis represents each decoder layer. RetroLM effectively retrieves crucial KVs.}
\label{fig:attn_map}
\vspace{-10pt}
\end{figure}

\subsection{GPU Memory Consumption}
As shown in Table \ref{tab:RULER} (Bottom), RetroLM significantly reduces memory consumption compared to full attention as input length increases. The memory usage of full attention grows quadratically with sequence length, while even flash attention exhibits linear growth. In contrast, RetroLM leverages streaming encoding with the page retriever to maintain a fixed KV budget during both the prefilling and decoding. This approach effectively minimizes peak memory consumption and implicitly compresses the context.

\subsection{Case Study}
\label{Sec:Case Study}
To further evaluate the effectiveness of KV cache level retrieval augmentation in RetroLM, we conduct case study using the MusiQue dataset \citep{bai2023longbench}, a challenging multi-hop QA task involving lengthy texts. As illustrated in Figure \ref{fig:attn_map}, we compare the full attention scores with those of the page retriever. Full attention fails to attend to the KVs containing the correct answer, resulting in an incorrect prediction. In contrast, our proposed page retriever effectively identifies and retrieves the relevant pages. Especially in the intermediate layers, page retriever demonstrates strong ability to focus on crucial KVs. Due to space constraint, more cases are presented in Appendix \ref{Appendix: Further Case Study}.

\subsection{Ablation Study}
\label{Sec:Ablation}
\textbf{Effectiveness of Page Retriever.} As presented in Table \ref{tab:analytical_experiment} (Top), to assess the effectiveness of page retriever training (Stage1), we implement the algorithmic framework of RetroLM without training the page retriever (w/o Stage1). The resulting test performance exhibits a 6.9 points degradation, underscoring the critical importance of training the page retriever for KV cache level retrieval augmentation.

\noindent \textbf{Effectiveness of Post Training.} As presented in Table \ref{tab:analytical_experiment} (Top), to assess the effectiveness of post-training (Stage2), we use the same unsupervised data to perform full-attention fine-tuning and evaluating on the Mistral model (Mistral-Finetuned). We then apply InfLLM \citep{infllm} algorithm using this model (InfLLM-Finetuned). While these approaches yielded modest performance improvements, they were markedly inferior to the results achieved by RetroLM after Stage-2 training. This demonstrates the necessity of adapting the model to sparse KV cache for effective KV cache usage and enhanced performance.

\noindent \textbf{Varying KV Budgets.} We assess the effectiveness of RetroLM under varying KV budgets. Using the RetroLM-stage1 model, which only trains the retriever module, we vray the token budget from 512 to 2048 and evaluate on the Longbench QA datasets. For comparative analysis, we include results from SnapKV with 1024 and 2048 token budgets and the full-attention model. The results are reported in Table \ref{tab:analytical_experiment} (Bottom).

Even with 512-token budget, RetroLM achieves an average score of 32.3 across the LongBench QA tasks, closely aligns with both SnapKV using a 2048-token budget and the full-attention model. 
As the token budget increases, we observe a clear trend of performance improvement. The significant performance gains on complex datasets like 2WikiMQA (+8.6) and HotpotQA (+6.7) suggest the effectiveness of RetroLM in complex long-context reasoning scenarios that demand robust information seeking and aggregation capabilities. 

\section{Conclusion}
In this paper, we introduce \textbf{RetroLM}, a novel RAG framework that enhances the performance of long-context processing by conducting retrieval augmentation at the KV cache level. Unlike traditional RAG methods that operate on raw tokens, RetroLM partitions the KV cache into contiguous pages and selectively retrieves the most crucial ones. To achieve precise and effective retrieval, we propose a specialized \textbf{page retriever} that evaluates page importance via fine-grained KV interactions. Additionally, we employ \textbf{post-training} on unlabeled data, enabling LLMs to better utilize retrieved KVs and improving end-to-end performance. Extensive evaluations are conducted on several standard long-context benchmarks.

\clearpage 



\bibliography{custom,anthology}
\bibliographystyle{acl_natbib} 

\clearpage
\appendix

\section{Training Detail and Data Formulation}
\label{Appendix: Training Detail and Data Formulation}
\subsection{Data Formulation}
\begin{figure*}[h]
\centering
\vspace{-35pt}
\includegraphics[width=0.98\linewidth]
{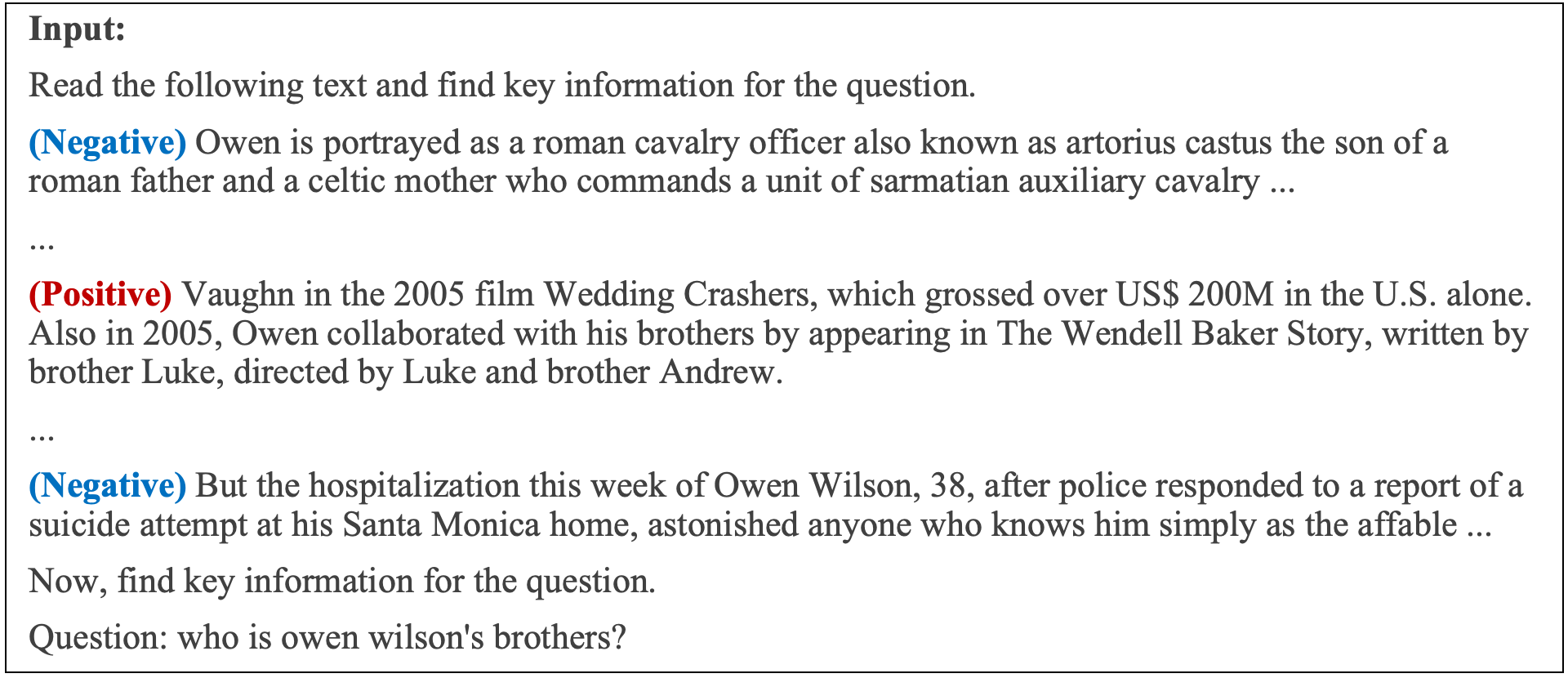}
\vspace{-8pt}
\caption{Example of weakly supervised data from msmarco \citep{bajaj2016ms} for stage-1 training. We concatenate the positive passage with hard negative passages in a random order, forming pseudo-texts up to a length of 8K tokens, and append the query to the end of the sequence. \textbf{ The page retriever is trained to identify useful KVs (positive passage) across each decoder layer via contrastive learning.}}
\label{fig:data_stage1_msm}
\end{figure*}

In this section, we introduce the detailed data curating method for the training of RetroLM. For stage-1 training, we exclusively train the KV retriever while keeping the backbone LLM fixed. We utilize a dataset of 50K pairwise examples sourced from MS MARCO \citep{bajaj2016ms}, formatted as illustrated in Figure \ref{fig:data_stage1_msm}. Each dataset entry comprises a web search query, along with 40 hard negative passages mined by BGE \citep{xiao2023c}. These passages are randomly interspersed with a positive passage, resulting in pseudo input texts up to 8,000 tokens in length. The query is appended at the end of the text, prompting the model to locate pertinent information corresponding to the input query. Leveraging the robust semantic understanding of the fixed backbone LLM, we employ contrastive learning across each decoder layer using bookmark tokens. This task requires the KV retriever to discern key KVs amidst substantial distractions. The input constraint of 8K tokens ensures high training efficiency and aims to establish the retrieval proficiency of the KV retriever.

\begin{figure*}[h]
\centering
\vspace{-75pt}
\includegraphics[width=0.98\linewidth]{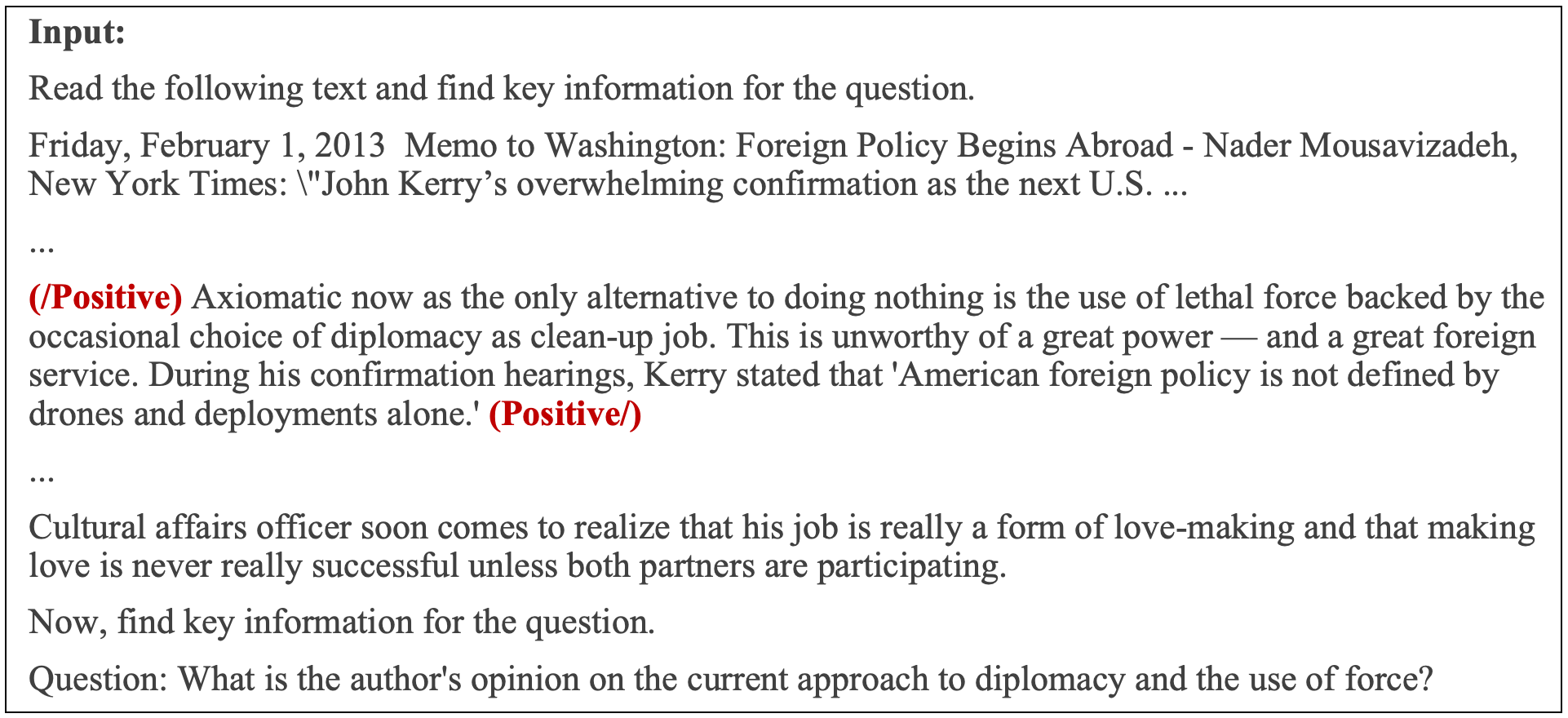}
\vspace{-8pt}
\caption{Example of synthetic data for stage-1 training.}
\label{fig:data_stage1_syn}
\end{figure*}

\begin{figure*}[t]
\centering
\vspace{-365pt}
\includegraphics[width=0.98\linewidth]{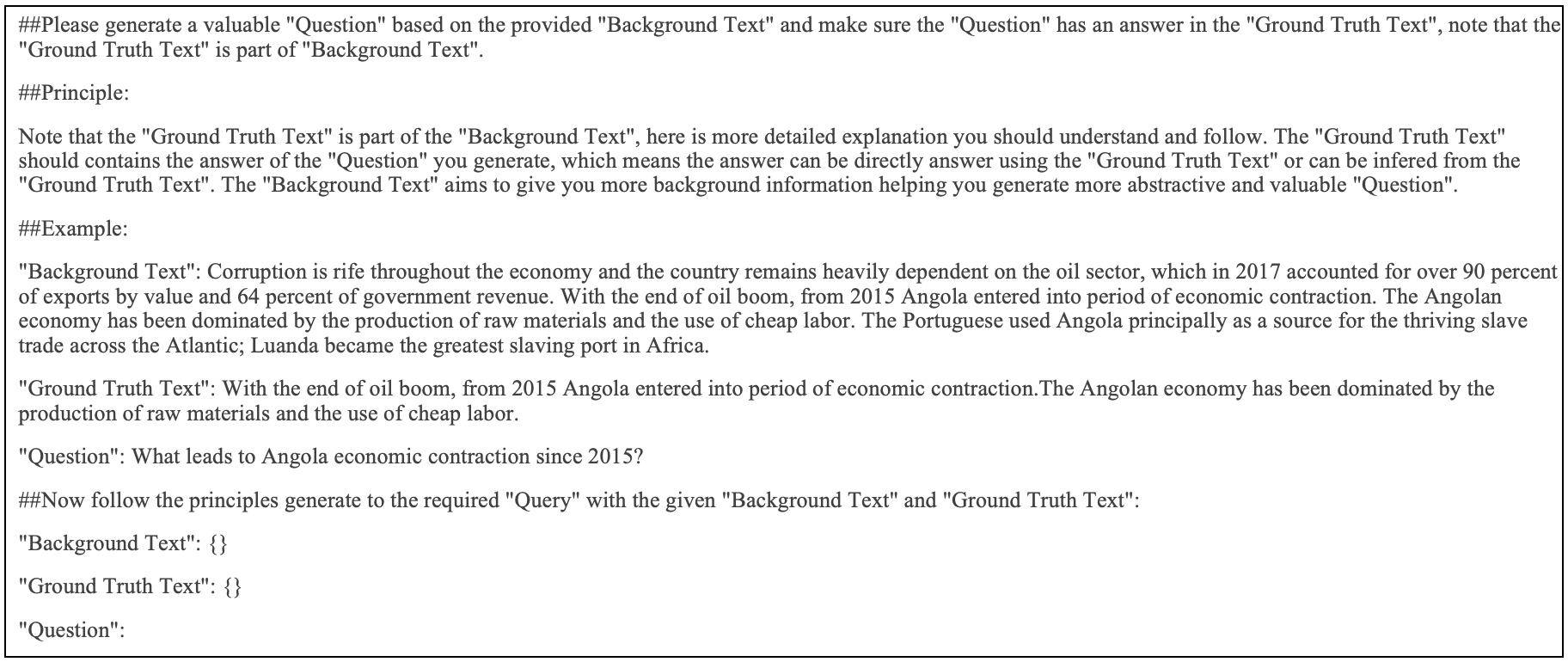}
\vspace{-8pt}
\caption{Prompt for construct question from Slimpagama \citep{shen2023slimpajama}.}
\label{fig:prompt_stage1}
\end{figure*}

Additionally, we generate 5K synthetic pairwise samples using text from Slimpajama \citep{shen2023slimpajama}, structured as depicted in Figure \ref{fig:data_stage1_syn}. Unlike discrete text spans from MS MARCO, these input texts consist of coherent passages, tasking the KV retriever with identifying key KVs for the query.
For the detailed curation method, we begin by sampling lengthy documents from Slimpajama. From these documents, we extract a segment (e.g., 100 words) as the \textit{Background Text}, and randomly select consecutive 1-5 sentences from this segment as the \textit{Ground Truth Text}. Employing the GPT API\footnote{https://platform.openai.com/}, we pose questions about the \textit{Background Text}, stipulating that the answers must be contained within the \textit{Ground Truth Text}. This method ensures that synthetic questions are contextually rich while ensuring that their answers remain within smaller semantic units. The prompt for constructing synthetic data are provided in Figure \ref{fig:prompt_stage1}. To maintain the quality of synthetic data, we ask ChatGPT to generate precise and insightful questions. If the generated text lacks meaningful information, it undergoes careful scrutiny and filtration.

For stage-2 training, we use 10K unsupervised text data from Slimpajama \citep{shen2023slimpajama}, with input lengths constrained to 12K tokens. The primary goal of stage-2 training is not to expand the context window but to enable RetroLM to adapt to a sparse KV cache and a streaming encoding paradigm. The corresponding ablation study is detailed in Sec. \ref{Sec:Ablation}.

\subsection{Implementation}
In this Section, we introduce the training and implementation details of RetroLM. We train RetroLM using \textbf{Mistral-7B-Instruct} and \textbf{Llama-3-8B-Instruct} as backbone respectively. During training, we set the page size for RetroLM to 128 tokens. This means the input text is divided into segments of 128 tokens, each appended with a bookmark token. It is important to note that the page size used during inference does not need to match the training page size; users can define it at any desired granularity. 

All the experiments take place on 8xA800 (80GB) GPUs. The learning rate for stage-1 training is 5×$10^{-6}$ and for stage-2 is 1×$10^{-6}$, the weight decay is 1×$10^{-2}$. The batch size is 1, where we accumulate the gradient over 16 steps. We leverage Flash-attention-v2 \citep{dao2023flashattention}, Gradient Checkpointing \citep{chen2016training}, and Deepspeed-Zero \citep{rajbhandari2020zero} to speed up the training. Throughout training, the peak CUDA memory usage is observed at about 40GB, which is attributable to the limited input length of 12K tokens.

\section{Further Case Study}
\label{Appendix: Further Case Study}
\begin{figure*}[t]
\centering
\vspace{-120pt}
\resizebox{1.01\linewidth}{!}{ 
    \includegraphics{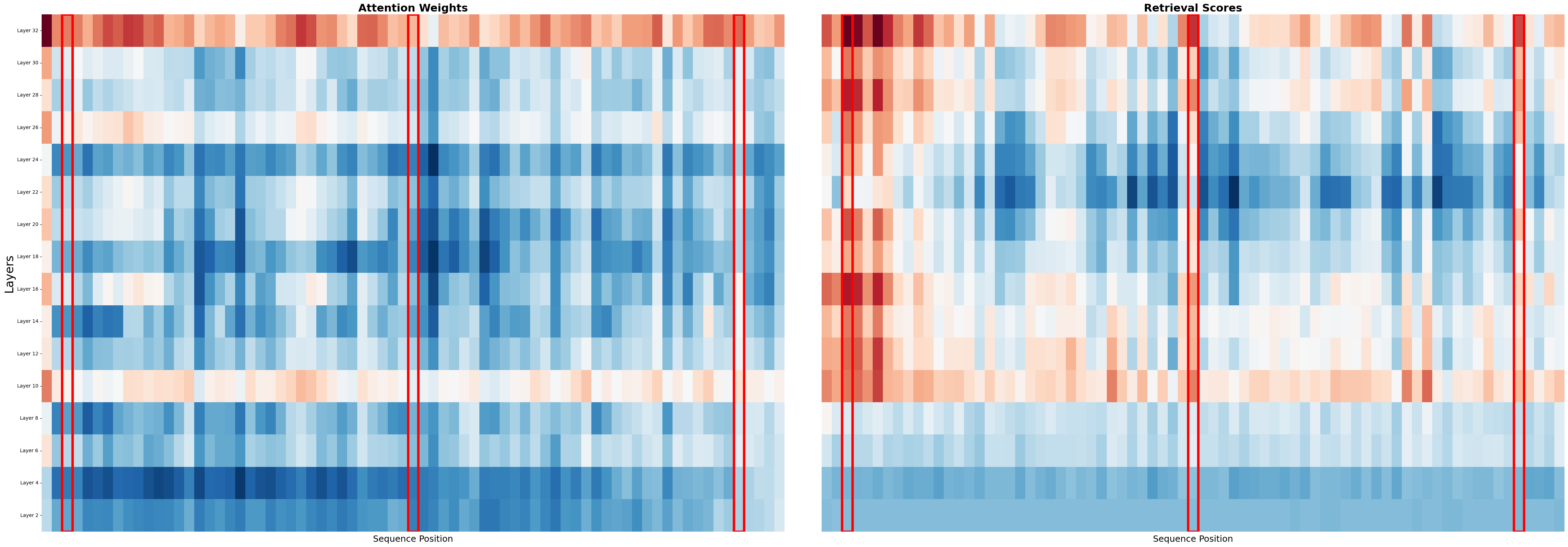}
}
\vspace{-18pt}
\caption{\small Case 2: attention score maps. Left: from original full attention. Right: score from RetroLM's KV retriever. \raisebox{0pt}{\scalebox{0.9}{\fcolorbox{red}{white}{Red squares}}} are answers for the multi-hop question. X-axis represents sequence position, Y-axis represents each decoder layer. RetroLM effectively retrieves crucial KVs.}
\label{fig:attn_map_2}
\vspace{-8pt}
\end{figure*}

\begin{figure*}[t]
\centering
\vspace{-220pt}
\resizebox{1.01\linewidth}{!}{ 
    \includegraphics{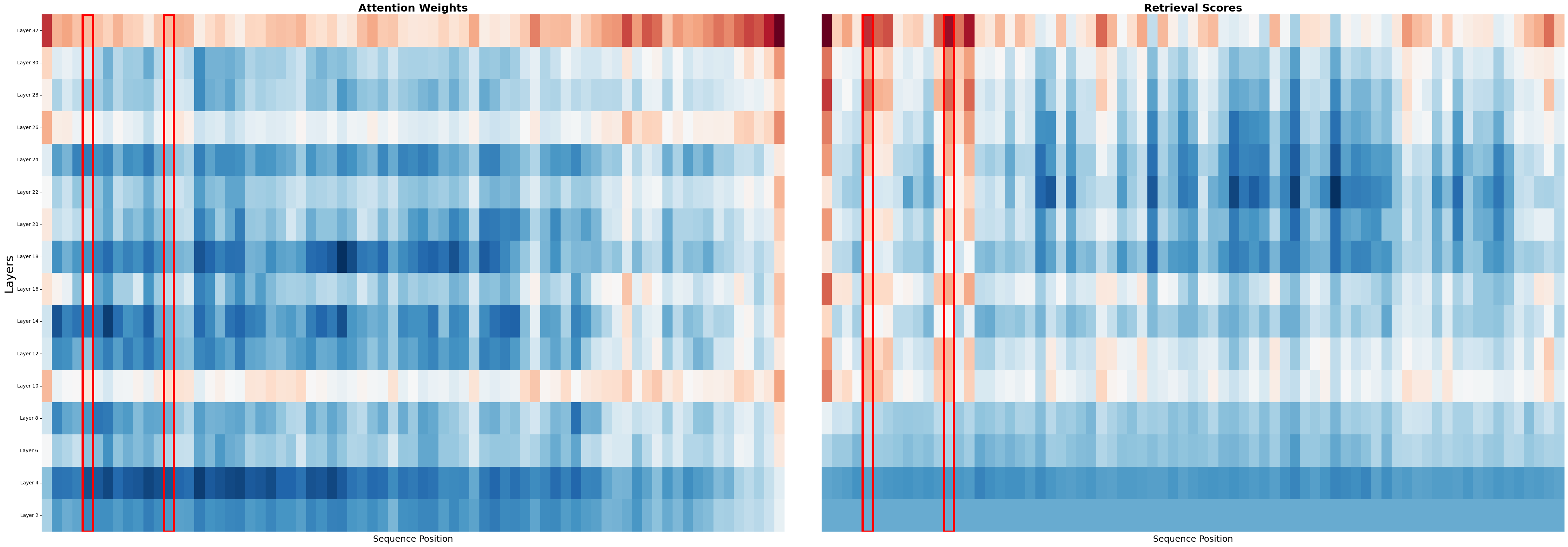}
}
\vspace{-18pt}
\caption{\small Case 3: attention score maps. Left: from original full attention. Right: score from RetroLM's KV retriever. \raisebox{0pt}{\scalebox{0.9}{\fcolorbox{red}{white}{Red squares}}} are answers for the multi-hop question. X-axis represents sequence position, Y-axis represents each decoder layer. RetroLM effectively retrieves crucial KVs.}
\label{fig:attn_map_3}
\vspace{-8pt}
\end{figure*}

In this section, we present more cases (Figure \ref{fig:attn_map_2} to Figure \ref{fig:attn_map_3}) to evaluate the effectiveness of KV cache level retrieval augmentation in RetroLM, using the MusiQue dataset \citep{bai2023longbench}. 

\clearpage

\clearpage

\end{document}